\pdfoutput=1

\documentclass[11pt,a4paper]{article}
\usepackage[final]{EACL2023}
\usepackage{times}
\usepackage{latexsym}

\usepackage{hyperref}
\usepackage{url}
\usepackage{breakurl}

\usepackage{graphicx}
\usepackage[export]{adjustbox}
\usepackage{bm}
\usepackage{here}
\usepackage{amsmath}
\usepackage{amssymb}
\usepackage{amsfonts}
\usepackage{multirow}
\usepackage{arydshln}
\usepackage{enumitem}
\usepackage{dsfont}

\usepackage{algorithm}
\usepackage{algpseudocode}
\usepackage{soul}
\usepackage{float}
\usepackage{subfigure}
\usepackage{microtype}

\title{Enhancing Performance on Seen and Unseen Dialogue Scenarios using Retrieval-Augmented End-to-End Task-Oriented System}

\author{ 
  \textbf{Jianguo Zhang}$^1$\footnotemark[1]~~~~ \textbf{Stephen Roller}$^2$ ~~~~ \textbf{Kun Qian}$^3$~~~~ \textbf{Zhiwei Liu}$^1$~~~~  \textbf{Rui Meng}$^1$ \\ \textbf{Shelby Heinecke}$^1$~~~~ \textbf{Huan Wang}$^1$~~~~\textbf{Silvio Savarese}$^1$ ~~~~\textbf{Caiming Xiong}$^1$ \\
  $^1$Salesforce AI  ~~~~ $^2$Character.AI ~~~~ $^3$Columbia University\\
 {\texttt{jianguozhang@salesforce.com}}
}

\date{}

\begin{document}
\maketitle

\footnotetext[1]{This work was partially conducted  during Jianguo's internship and Stephen's full-time employment at Meta AI Research (FAIR).}

\begin{abstract}
End-to-end task-oriented dialogue (TOD) systems have achieved promising performance by leveraging sophisticated natural language understanding and natural language generation capabilities of pre-trained models. This work enables the TOD systems with more flexibility through a simple cache. 
The cache provides the flexibility to dynamically update the TOD systems and handle both existing and unseen dialogue scenarios. 
Towards this end, we first fine-tune a retrieval module to effectively retrieve the most relevant information entries from the cache. We then train end-to-end TOD models that can refer to and ground on both dialogue history and retrieved information during TOD generation. The cache is straightforward to construct, and the backbone models of TOD systems are compatible with existing pre-trained generative models. Extensive experiments demonstrate the superior performance of our framework, with a notable improvement in non-empty joint goal accuracy by $6.7\%$ compared to strong baselines.


\end{abstract}

\section{Introduction}




Task-oriented dialogue (TOD) systems play an important role in various applications, such as restaurant booking, alarm setting, and recommendations~\cite{gao2018neural,xie2022converse}. These systems can be broadly categorized into two groups: pipeline-based dialogue systems and end-to-end dialogue systems. 
Pipeline-based dialogue systems consist of four separate modules: a natural language understanding (NLU) module to detect user intents, a dialogue state tracking (DST) module to track user belief states across dialogue turns, a dialogue management (DM) module to decide system actions based on dialogue states, and a natural language generation (NLG) module to generate natural-language responses.
However, the pipeline-based approach is annotation-intensive, prone to error propagation, and challenging to scale~\cite{hosseini2020simple,zhang2020find,feng2023fantastic}.

\begin{figure}
	\begin{center}
    	\includegraphics[width=1.0\linewidth]{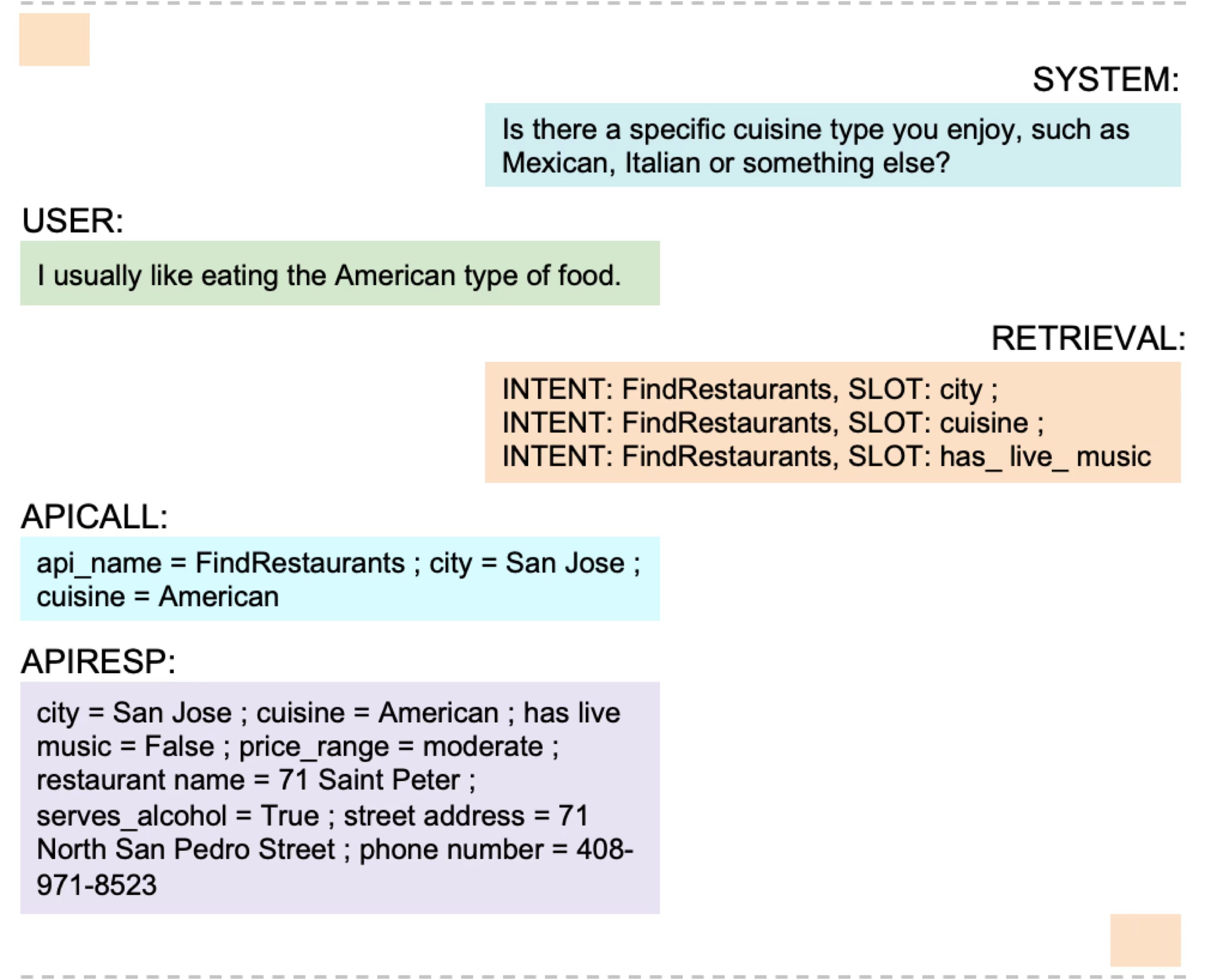} 
    \end{center}
\caption{An example of the auto-regressive TOD with retrieved slot information from cache. The APICALL generation process is shown, with $N$ set to 3 for the retrieval module.
}
\label{fig:example_tod}
\end{figure}

Recently, various approaches have been proposed to utilize sequence-to-sequence models for generating dialogue states and responses in an end-to-end manner~\cite {ham2020end,lin2020mintl,yang2020ubar,gao2021hyknow,chen2021teaching,peng2021soloist,liu2021pretraining,he2022galaxy,feng2023fantastic}. 
Compared with the pipeline-based systems, these approaches demonstrate effectiveness on public datasets requiring fewer direct annotations, such as user intents and dialogue acts. Additionally, they leverage the capabilities of large-scale pre-trained language models, such as GPT-2~\cite{radford2019language}, T5~\cite{raffel2019exploring}, and BART~\cite{lewis2020bart}, for improved performance in NLU and NLG tasks. 
However, these approaches are limited in their ability to dynamically handle both existing and emerging intents and slots,
particularly in the context of unseen dialogue scenarios such as new domains and services~\cite{hosseini2020simple,peng2021soloist,rastogi2020schema}. 

Simultaneously, research on open-domain question answering and dialogue systems has explored the use of retrieval-augmented models. These models retrieve relevant information from a passage, database, APIs, etc., and incorporate it into the generation process, improving answer quality or dialogue responses~\cite{karpukhin2020dense, izacard2021leveraging, dinan2018wizard, lewis2020retrieval, shuster2021retrieval}. Inspired by these ideas, we combine both worlds and propose an end-to-end TOD framework with a retrieval system that addresses the challenge of handling both existing and zero-shot unseen dialogue scenarios.

Our approach involves training the end-to-end TOD models with a cache that contains accessible domains, intents, slots, and APIs. The cache can be constructed based on the schema or database, or by extracting information from accessible dialogues when the schema or database is not fully accessible. 
The cache serves as a reference point, allowing the models to ground their responses in the retrieved information. 
By incorporating a retrieval module and leveraging this cache of knowledge, our system enhances the flexibility and adaptability to handle both existing and unseen intents and slots. It also ensures robust performance in novel dialogue domains and services where the model hasn't been explicitly trained.
Figure~\ref{fig:example_tod} shows an illustrative example of our approach, demonstrating how the RETRIEVAL module retrieves relevant information, such as slots in this case, from the cache to enrich the system's understanding and generate more accurate responses. The APICALL represents the dialogue states from the system side, and APIRESP returns information from external API interactions between the system and system databases.

To build an accurate end-to-end TOD system with the benefits of a simple cache, 
we fine-tune a retrieval module to effectively retrieve the most relevant and informative information from the cache, using a Top-$N$ retrieval strategy. Then we integrate the retrieval module into the generative model to facilitate end-to-end TOD generation.
We evaluate our approach on the publicly available Google Schema-Guided Dialogue dataset (SGD)~\cite{google-schema-guided}, which includes a significant number of unseen dialogue domains and services in the development and test sets.

The contributions of this paper are as follows: 
(1) We design a simple yet effective end-to-end TOD framework with a cache that enables dynamic handling of intents and slots. The framework is compatible with existing pre-trained generative models, and enhances the system's robustness. 
(2) We provide experimental results that demonstrate the superior performance of our approach compared to strong baselines. 
It achieves $6.7\%$ improvement in non-empty joint goal accuracy, demonstrating the effectiveness in handling various dialogue scenarios, including the challenging zero-shot unseen dialogues.
(3) 
We conduct comprehensive ablation studies and analyses to provide insights into the impact of different components and design choices within our framework. 



\section{Related Work}


\paragraph{End-to-End TOD Systems}
End-to-end TOD models have shown promising performance on public dataset~\cite{ham2020end,lin2020mintl,yang2020ubar,gao2021hyknow,chen2021teaching,peng2021soloist,liu2021pretraining,he2022unified,he2022galaxy,feng2023fantastic,bang2023task}. 
These approaches typically follow common patterns:
(1) Rely on powerful pre-trained seq2seq models. (2) Use language modeling objectives to generate NLU and NLG outputs, sometimes augmented with auxiliary multi-task goals like DST loss. (3) Either fine-tune models directly on the target dataset or conduct pre-training on multiple TOD dialogue datasets. (4) Employ data augmentation techniques such as back-translation and entity replacement due to the challenges in collecting large-scale TOD corpora. 
For example, \newcite{hosseini2020simple} fine-tunes DistilGPT2 for TOD. The model generates user belief states and system responses in an auto-regressive way. 
\newcite{peng2021soloist} introduces two auxiliary tasks for belief state prediction and grounded response generation and pre-train language models first on multitple TOD dataset.  
\newcite{gao2021hyknow} enables the belief state to interact with both structured and unstructured knowledge. 
\newcite{feng2023fantastic} designs a reward-function learning objective to guide the model's generation.
While these methods have demonstrated effectiveness on public datasets, they have limitations in handling unseen dialogue scenarios such as unseen domains and services.

\paragraph{Retrieval-Augmented Models}
Retrieval augmented approaches have been widely used in open-domain question answering. For instance,
\newcite{karpukhin2020dense} proposes a BERT-based~\cite{bert} dual-encoder framework to retrieve passages from Wikipedia, which is  further incorporated into open-domain conversations to reduce hallucination and enrich engagement with users~\cite{shuster2021retrieval, komeili2021internet}. 
These models retrieve information related to the query from a knowledge base of sentences and ground the generation response on this information \cite{dinan2018wizard,lewis2020retrieval}. 
Inspired by these works, we explore the integration of retrieval modules into end-to-end TOD systems, leveraging the retrieval-augmented approach to enhance the system's performance in handling both existing and novel dialogue scenarios.




\section{TOD Systems with a Simple Cache}

We present an end-to-end transformer-based framework with a simple cache that is compatible with multiple generative models, including BART, T5, GPT2, etc. Our framework enables dynamic handling of intents, slots, and APIs while maintaining flexibility in choosing the backbone model.

Generally, our framework consists of two  parts: a retrieval model for retrieving the most relevant and informative information from the cache, and an end-to-end TOD model that generates APICALLs and system responses based on the dialogue history and the retrieved information. The retrieval model functions by retrieving intents, slots, APIs, and other relevant information from the cache.

\begin{figure*}
	\begin{center}
    	\includegraphics[width=\linewidth]{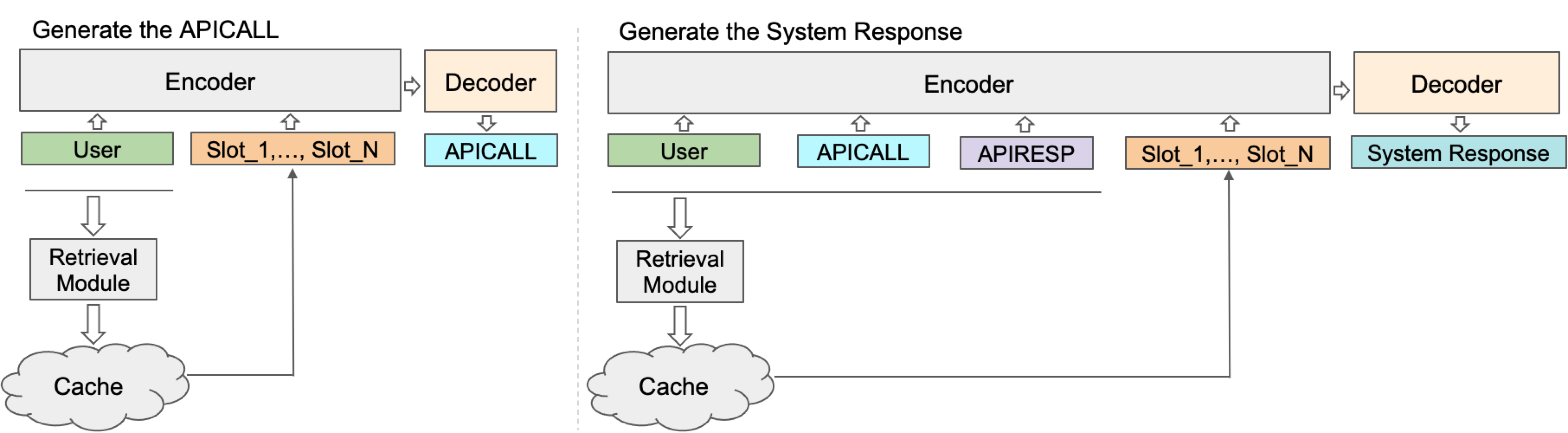}
    \end{center}
\caption{Illustration of the end-to-end framework with a simple cache. 
The left figure illustrates the generation of an APICALL. In this process, the retrieval module extracts relevant information, like slots, from the cache. The decoder then uses the retrieved information, combined with the dialogue history, to generate the APICALL.
The right figure depicts the continuation of the dialogue, generating the system response. The system retrieves additional information from the cache and incorporates all previous information to generate a system response. 
By decoupling the generation of APICALLs from system responses, we aim to clearly represent the framework's components and their interactions in an end-to-end setting.
}
\label{fig:tod_framework}
\end{figure*}

Figure~\ref{fig:tod_framework} illustrates one simple variant of our framework, which is an encoder-decoder architecture. In this variant, the retrieved information such as slots are stacked together. We also introduce another variant in Sec.~\ref{sec:labels}, where each retrieved information is concatenated with the dialogue history and then all the information are concatenated together before being sent to the decoder. 



\begin{table*}[]
\resizebox{\linewidth}{!}{
\begin{tabular}{ll}
\hline
Cache Templates                                                                                        & Examples                                                                                                                                                       \\ \hline
INTENT: intent name, SLOT: slot name                                                                                  & INTENT: findrestaurants, SLOT: city                                                                                                                                          \\ \hdashline
\begin{tabular}[c]{@{}l@{}}intent name, slot name, service description, \\  intent description, slot description\end{tabular} & \begin{tabular}[c]{@{}l@{}}findrestaurants, city, a leading provider for restaurant search and reservations, \\  find a restaurant of a particular cuisine in a city, city in which the restaurant is located\end{tabular} \\ \hdashline
API-information                                                                                  & \begin{tabular}[c]{@{}l@{}}api\_name = FindRestaurants; optArg = has\_live\_music, price\_range, \\ serves\_alcohol; reqArg = city, cuisine\end{tabular}       \\ \hline
\end{tabular}}\caption{Several typical templates of the simple cache construction, where each template represents one type of cache. 
Some other templates can be found in Table \ref{tab:DPR}.}\label{table:cache}
\end{table*}

\subsection{Construction of Cache}\label{section-cache}
In this section, we describe the construction of a simple cache that provides necessary information for the model's referencing and grounding procedure. The cache consists of intents, slots, and APIs extracted from the schema and database. In cases where the schema or database is not fully accessible, we extract information from accessible dialogues.
During training, it is important to note that the cache exclusively contains information relevant to the training dialogues and does not incorporate any unseen information of dialogues in the test set.

Since there are different ways to construct a cache, we design various templates to formalize the retrieved information. Table~\ref{table:cache} presents several templates that we utilize. 
One example is the ``\textit{API-information}'' template, where an API includes all the intents and relevant slots mentioned throughout the whole dialogue. Although this template may contain redundant information as some intents and slots may not be mentioned initially, it allows us to evaluate the model's ability to disregard irrelevant details.

In addition to the listed templates, we explore several other templates with special tokens such as ``\textit{[INTENT] intent name [SLOT] slot name}'', as well as different orderings of intents and slots, such as ``\textit{intent name, intent description, slot name, slot description}'' and ``\textit{intent name, slot name, intent description, slot description}''. We conduct an in-depth analysis of the effects of different cache templates in the experimental section. 

\subsection{Retrieval Module}
After constructing the cache, we fine-tune a retrieval model to effectively retrieve the most relevant and informative information for the dialogue context. 
Given a dialogue history $c$, the TOD system utilizes a retrieval module to retrieve Top-$N$ most relevant information $s_1,\ldots, s_N$ from the cache. Firstly, based on the dialogue history,  the system triggers the retrieval module to generate an APICALL, which includes relevant mentioned intents, slots and values. Subsequently, the system continues to use the retrieval module to generate a system response based on all previous information. 

To ensure accurate retrieval from the cache, we fine-tune a dense passage retriever (DPR) model~\citep{karpukhin2020dense}, which is a BERT-based dual-encoder framework optimized via contrastive learning. Specifically, we obtain the hidden representation $\mathbf{h}_c$ for the dialogue history using an encoder model, \textit{e.g.}, $\mathbf{h}_c=\text{BERT}_c(c)$. Similarly, we use another BERT encoder to obtain the feature representation $\mathbf{h}_{s}$ for each retrieved information entry from the cache, \textit{i.e.}, $\mathbf{h}_{s}=\text{BERT}_s(s)$. The similarity between the dialogue history and the retrieved information entry is: $sim(c, s)=\mathbf{h}_c^{T}\odot\mathbf{h}_{s}$.

For each dialogue history, there are $n$ relevant (positive) entries and $m$ irrelevant (negative) entries, where $n$ and $m$ may vary as each dialogue history would contain different active intents and slots. 
Our objective is to learn a function that minimizes the distance between pairs of relevant dialogue histories and information entries than the irrelevant pairs. The corresponding loss function for a specific pair is as follows:
\begin{equation}
    \label{eq:hard-negative-eq}
    \small
    \mathcal{L}_{\text{api}}(c,s_{1}^{+},s_{1}^{-},\ldots, s_{m}^{-})=-\log\frac{\exp(\text{sim}(\mathbf{h}_c,\mathbf{h}_{s_{1}^{+}}))}{\sum_{j=1}^{m}\exp(\text{sim}(\mathbf{h}_c,\mathbf{h}_{s_{j}^{-}}))}\,.
\end{equation}

Once the retrieval module is fine-tuned, it is  incorporated into the end-to-end sequence-to-sequence task-oriented dialogue generative model. The parameters of the retrieval module remain fixed during training of the generative model.


\paragraph{Negative Sampling} 
In the training process, we employ negative sampling to include retrieved information entries that are irrelevant to the dialogue history. We utilize both natural and hard negative pairs to enhance the robustness and performance of the retrieval module.

For natural negative pairs, we consider pairs such as ``\textit{irrelevant intent, irrelevant slots}'' as counterparts to the positive pairs of ``\textit{relevant intent, relevant slots}''. Additionally, we construct hard negative pairs that pose a more challenge to the retrieval module. These hard negative pairs include combinations such as ``\textit{relevant intent, irrelevant slots from the same relevant intent}'' and ``\textit{irrelevant intents that are semantically similar to the relevant intent, along with relevant slots from the relevant intent}''. By incorporating these hard negative pairs, we encourage the retrieval module to learn to differentiate between relevant and irrelevant information effectively.


\subsection{End-to-End TOD Systems}
Our end-to-end TOD framework generates the APICALL and system response in an auto-regressive manner. Figure~\ref{fig:example_tod} provides an example of this process. The APICALL represents the dialogue states from the system side, and same with previous work~\citep{hosseini2020simple,peng2021soloist}, it is an intermediate step of the system response generation, and they share the same model framework to generate tokens autoregressively.    

For each dialogue turn, the TOD framework triggers the retrieval module twice. The system first retrieves the Top-$N$ information entries from the constructed cache, \textit{i.e.},
\begin{equation}
    \small
\text{Top-$N$ info} = \text{Retrieval}(c)\,.
\end{equation}
Then it generates an APICALL using the retrieved information, \textit{i.e.}, 
\begin{equation}
    \small
\text{APICALL} = \text{TOD}(c, \text{Top-$N$ info)}\,.
\end{equation}
After that the TOD framework retrieves another set of Top-$N$ information entries from the cache, considering the generated APICALL, \textit{i.e.}, 
\begin{equation}
    \small
\text{Top-$N$ info} = \text{Retrieval}(c, \text{APICALL}, \text{APIRESP})\,,
\end{equation}
where APIRESP is automatically obtained from corresponding API, without the need for prediction.

Finally, the system generates a system response using the following inputs:
\begin{equation}
\small
    \text{Response} = \text{TOD($c$, APICALL, APIRESP, Top-$N$ slots)}\,.
\end{equation}

\section{Experimental Settings}
\subsection{Dataset}
 A substantial number of end-to-end TOD works~\citep{hosseini2020simple,peng2021soloist,lin2020mintl,yang2020ubar,su2021multi,he2022galaxy,feng2023fantastic} commonly employ the MultiWOZ datasets~\citep{budzianowski2018multiwoz,zang2020multiwoz}. However, these studies primarily focus on full-shot and few-shot learning, 
 and the scope for zero-shot evaluation appears somewhat constrained given that MultiWOZ only has five domains and approximately $35$ slots, all of them are presented in the training set. In contrast, our work aims to assess the system across large-scale multi-domain dialogue scenarios. We utilize the  Google Schema-Guided Dialogue (SGD) dataset~\cite{rastogi2020towards}. \footnote{\scriptsize \href{https://github.com/salesforce/DialogStudio/tree/main}{SGD processed dataset.}} SGD provides a more expansive dialogue landscape with over 16,000 multi-domain conversations that span more than 16 domains, 26 services, and 200 slots. Notably, the SGD includes a significant number of unseen domains/APIs in both the development and test sets, thereby increasing the complexity and diversity of the dataset. A substantial 45\% of the dialogue turns in the development set, and a significant 77\% of the dialogue turns in the test set, incorporate at least one service not included in the training set.
Table~\ref{dataset:SGD} summarizes the statistics of SGD.  
 

\begin{table}[]
\resizebox{\linewidth}{!}{
\begin{tabular}{l|cccccc}
 \hline
            & Dialogues &Services & ZS Services & ZS Turns \\ \hline
Train       & 16142          & 26                      & -  & -                \\
Dev. & 2482         & 17                        & 8   & 45\%               \\
Test        & 4201         & 21                      & 11  & 77\%             \\  \hline
\end{tabular}}\caption{Data Statistics of SGD. ZS: Zero-Shot. }\label{dataset:SGD}
\end{table}

\begin{table*}[t]
\centering
\resizebox{1.0\linewidth}{!}{
\begin{tabular}{l|cccccc}
\hline
        & PPL            & Overall JGA                  & Non-Empty JGA       & Token EM        & BLEU-4 \\ \hline
MinT (BART-Large) ~\citep{chen2021teaching}    & 2.385          & 0.812           & 0.364                 & 0.497          & \textbf{0.179} \\
T5DST ~\citep{lee2022sgd}    & 2.419          & 0.810           & 0.361                 & 0.491          & 0.170 \\
FiD-TOD & \textbf{2.133} & \textbf{0.829}       & \textbf{0.431} & \textbf{0.501} & \textbf{0.179} \\ \hline
\end{tabular}}\caption{Testing results on the SGD dataset. } \label{tab:tod-main_results}
\end{table*}

\begin{table*}[]
\centering
\resizebox{1.0\linewidth}{!}{
\begin{tabular}{l|ccccc}
\hline
Cache Templates                                                                                                                               & Top-1           & Top-2           & Top-3  & Top-4  & Top-5  \\ \hline
INTENT: intent name, SLOT: slot name                                                                                               & 0.833 & 0.882 & 0.914 & 0.945 & 0.960 \\ \hdashline
\begin{tabular}[c]{@{}l@{}}INTENT: intent name,  service description, \\ intent description, SLOT: slot name, slot description\end{tabular} & 0.887 & 0.922 & 0.952 & 0.976 & 0.980 \\ \hline
intent name, slot name, intent description, slot description                                                                                         & 0.835 & 0.906 & 0.928 & 0.946 & 0.955 \\ \hdashline
\begin{tabular}[c]{@{}l@{}}intent name, slot name, service description, \\ intent description, slot description\end{tabular}                         & 0.913 & 0.943 & 0.965 & 0.977 & 0.981 \\ \hline
API-information                                                                                                                      & 0.844 & 0.927 & 0.956 & 0.962 & 0.967 \\ \hline
\end{tabular}} \caption{Top-5 retrieval accuracy on the test set of SGD. } \label{tab:DPR}
\end{table*}

\subsection{Models}\label{sec:labels}


In term of baselines, we adopt  \cite{lin2020mintl,chen2021teaching} and implement their model MinTL (BART-Large).  We also implement T5DST from \cite{lee2022sgd}, which achieves strong performance on MultiWOZ 2.2~\citep{zang2020multiwoz}.  Since our end-to-end TOD framework is compatible with existing pre-trained generative models, we experiment with BART, GPT2 and T5. Interestingly, we found that BART-Large (406M) perform comparably with T5-Large (770M), despite having fewer parameters. Moreover, it outperformed many models developed by teams in DSTC8~\citep{rastogi2020schema}, where the majority of models are BERT-based classification models. Thus, we select BART-Large as our primary backbone model.

\begin{figure}
	\begin{center}
    	\includegraphics[width=\linewidth]{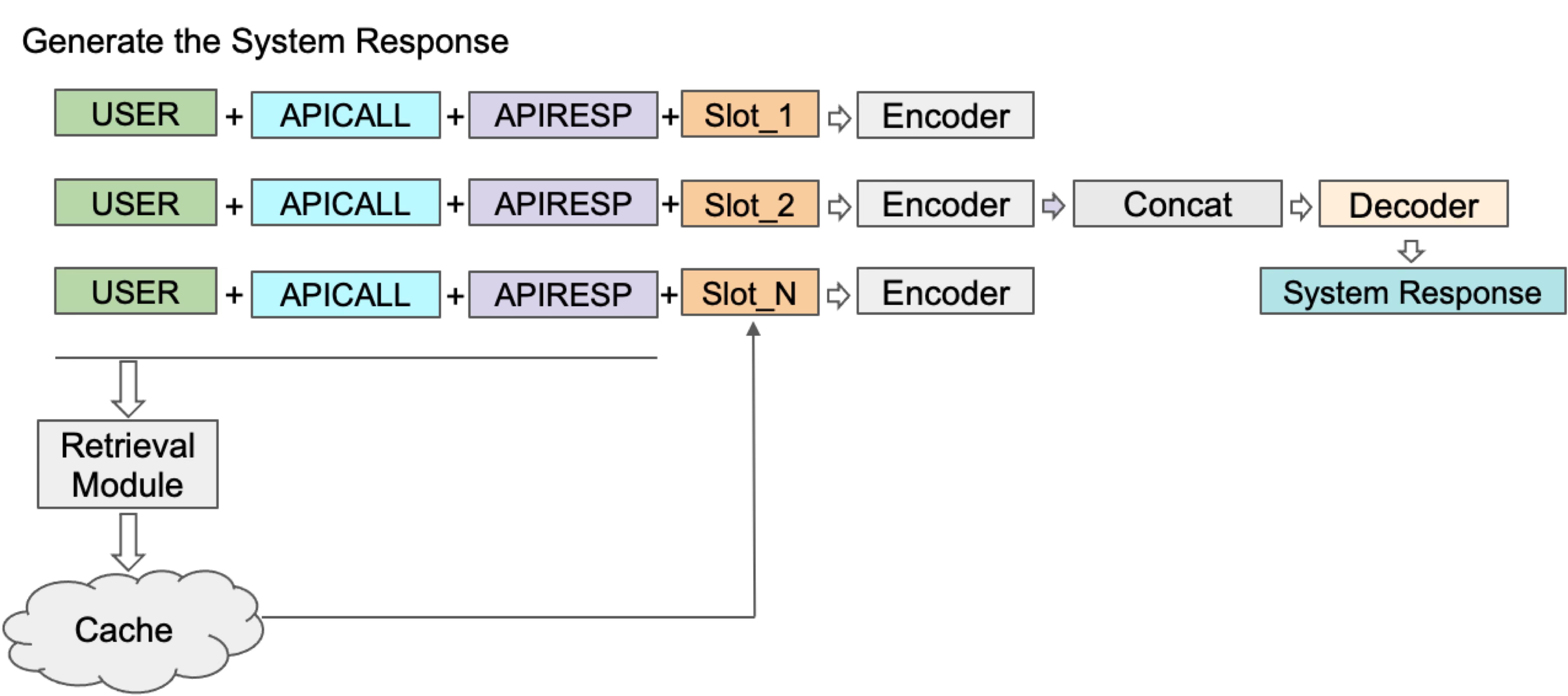}
    \end{center}
\caption{Illustration of FiD-TOD-NoStack framework.
}
\label{fig:tod_framework_2}
\end{figure}

Inspired by previous model designs in open-domain question answering~\citep{lewis2020retrieval,izacard2021leveraging}, we design two variants for end-to-end TOD systems. The first, named Fusion-in-Decoder TOD (FiD-TOD), is illustrated in Figure~\ref{fig:tod_framework}, In this model, the retrieved information such as slots, are stacked together. Notably, when the retrieval model is not incorporated, FiD-TOD becomes identical to MinTL (BART-Large).  The second variant FiD-TOD-NoStack, is depicted in Figure~\ref{fig:tod_framework_2} and is used as ablation study. In this model, the retrieved information is not directly stacked, and instead, the dialogue history is concatenated with each retrieved information entry and then sent to the shared encoder. 

Regarding the generative model, we truncate the tokens of dialogue history to 256, and retrieve Top-5 most relevant information entries from the cache, unless otherwise specified. For DPR fine-tuning, we align one hard negative pair to each positive pair.  We employ the preset hyperparameters from the ParLAI code, \footnote{\scriptsize \href{https://github.com/facebookresearch/ParlAI}{ParLAI platform.}}
 such as setting the learning rate to 5e-5, batch size to 32, etc. Initially, we conducted experiments with slight alterations in hyperparameters and observed no statistically significant difference on performance. We selected the best model based on its performance on the development set.

The retrieve model is fine-tuned up to 3 epochs based on open-sourced DPR~\citep{karpukhin2020dense}, and the generative model is fine-tuned up to 4 epochs with an overall batch size of $64$ on $8$ Nvidia Tesla V100 GPUs.
All experiments are based on public code from the ParLAI platform.

\subsection{Evaluation Metrics}

We evaluate the end-en-end TOD framework using the following ParLAI metrics: (1) Top-$N$ accuracy: It evaluates the retrieval module through checking whether the ground-truth slot appears in the Top-$N$ predicted candidates~\cite{karpukhin2020dense}. (2) Joint Goal Accuracy (Overall JGA): It evaluates whether the predicted APICALL on both seen and unseen services is correct or not, specifically. JGA is $1$ if the model correctly predicts all intent, slots and corresponding values in the APICALL. Otherwise, JGA is $0$. 
(3) Non-Empty JGA: It evaluates whether overall JGA is correct if the model calls the API on both seen and unseen scenarios. In SGD, most dialogue turns would not trigger an API retrieval, resulting in empty APICALLs, and identifying Empty JGA is relatively quite easy~\citep{chen2021teaching}. Moreover, most services, intents and slots in the test set are unseen. Therefore, we focus on Non-Empty APICALL turns and treat it as the most crucial metric for evaluating the model's performance on both seen scenarios and its zero-shot generalization ability on unseen scenarios.
(4) Token EM:  It evaluates the utterance-level token accuracy. Roughly corresponds to perfection under greedy search (generative only). (5) Perplexity (PPL): It measures the generative model’s ability to predict individual tokens. (6) BLEU-4: It measures the BLEU score~\cite{papineni2002bleu} between the predicted system response and the reference response.  

\begin{table*}[]
\centering
\resizebox{1.0\linewidth}{!}{
\begin{tabular}{l|cccccc}
\hline
                                                                                                                                          & PPL   & Overall JGA    &  Non-Empty JGA & Token EM & BLEU-4 \\ \hline
MinT (BART-Large) ~\citep{chen2021teaching}                                                                                                                              & 1.700 & 0.876     & 0.586          & 0.538   & \textbf{0.221} \\ \hline
INTENT: intent name, SLOT: slot name                                                                                                      & 1.688 & 0.889    & 0.633        & 0.538   & 0.212 \\ \hdashline
intent name, slot name, intent description, slot description                                                                              & 1.679 & 0.895    & \textbf{0.661}         & 0.541   & 0.215 \\ \hdashline
\begin{tabular}[c]{@{}l@{}}intent name, slot name, service description,\\ intent description, slot description\end{tabular}               & 1.679 & 0.894     & 0.649         & 0.541   & 0.212 \\ \hdashline
\begin{tabular}[c]{@{}l@{}}INTENT: intent name, service description,\\ intent description, SLOT: slot name, slot description\end{tabular} & \textbf{1.676} & \textbf{0.897}     & 0.660          & \textbf{0.545}   & 0.217 \\ \hline
\end{tabular}}\caption{Performance of FiD-TOD on the development set with variations of cache templates.}\label{table:variants-of-db}
 \vspace{-1.0em}
\end{table*}


\begin{table*}[]
\centering
\resizebox{0.98\linewidth}{!}{
\begin{tabular}{l|cccccc}
\hline
                                 & PPL   & Overall JGA    & Non-Empty & Token EM & BLEU-4 \\ \hline
MinT (BART-Large) ~\citep{chen2021teaching}                      & 1.700 & 0.876    & 0.586            & 0.538   & \textbf{0.221} \\ \hline
FiD-TOD w/ API-information   (N=1) & \textbf{1.653} & 0.896     & 0.658          & 0.543   & 0.218 \\ \hdashline
FiD-TOD w/ API-information   (N=5) & 1.655 & \textbf{0.897}    & \textbf{0.663}           & 0.544   & 0.219 \\ \hline
FiD-TOD-NoStack                & 1.683 & 0.895   & 0.653           & 0.543   & 0.215 \\ \hdashline
FiD-TOD                          & 1.676 & \textbf{0.897}   & 0.660           & \textbf{0.545}   & 0.217 \\ \hline
\end{tabular}}\caption{Results on development set. By default, retrieval module retrieves Top-5 information entries from cache.} \label{tab:tod-ablation}
 \vspace{-1.0em}
\end{table*}

\section{Experimental Results}\label{sec:exp-results}

\subsection{End-to-End TOD Performance}
Table \ref{tab:tod-main_results} shows the overall performance on the test set. FiD-TOD outperforms baselines across most metrics. Specifically, it improves the essential NLU metric, \textit{i.e.},  Non-Empty JGA, by  $6.7\%$. This demonstrates the model's enhanced capability in handling both seen dialogue scenarios and, notably, its capacity for zero-shot handling of unseen scenarios. As MinT (BART-Large) corresponds to the FiD-TOD without the retrieval model from the cache, this comparison highlights the significant benefits that our design brings to the handling of unseen dialogs. Additionally, the other metrics related to NLG are also slightly improved. 


\subsection{Retrieval Performance}

We hope the model can generalize well as there could be many new  intents and slots in real world. 
\paragraph{General} As shown in Table \ref{tab:DPR}, our model shows effective Top-5 retrieval accuracy on the test set, keeping in mind that more than half of the services and slots are unseen in this set. The model shows good Top-1 accuracy and above $96\%$ Top-5 accuracy, demonstrating strong abilities for handling  both seen and unseen intents and slots. Compared to only using names, adding related service and intent descriptions improves the Top-1 accuracy by more than $5\%$. This suggests that incorporating descriptions can enhance the model's ability to generalize to unseen dialogue scenarios. 

\paragraph{API-information} When evaluating the ``\textit{API-information}'', where a single API entry in the cache encompasses all intents and slots information for the whole dialogue. We see that the model has high Top-1 accuracy and Top-5 accuracy. This suggests that the model has a high potential to retrieve all the related intents and slots information with a single retrieval attempt. 

\paragraph{Orders and Special Tokens} We test with different templates, such as switching orders of intents and slots, and find no significant differences. We also find that adding the special tokens ``\textit{INTENT}'' and ``\textit{SLOT}'' slightly decreases the Top-1 accuracy. 

\paragraph{Negative Sampling} Experiments with both normal and hard negative pairs, including varying numbers of hard negative pairs, showed no significant impact on retrieval performance.  This could be attributed to the fact that, unlike longer passages in question answering, dialogue intents, slots, and APIs are generally easier to distinguish when they are referenced  in the dialogue context.

 
\subsection{Performance of Variants of Cache on End-to-End TOD}
As our design involves several templates for the cache, we aim to assess the impact of various cache templates on the performance of the end-to-end TOD system.
Table ~\ref{table:variants-of-db} shows that FiD-TOD using only names already outperforms MinT (BART-Large), and adding descriptions further improves the performance. For instance,  FiD-TOD with cache template ``\textit{INTENT: intent name, service description, intent description, SLOT: slot name, slot description}'' surpassess both MinT (BART-Large) and FiD-TOD with cache template ``\textit{INTENT: intent name, SLOT: slot name}'' by $7.4\%$ and $2.7\%$ in terms of Non-Empty JGA, respectively. 

\begin{table*}[]
\resizebox{\linewidth}{!}{
\begin{tabular}{l|l}
\hline
...                                         &  ...                                                                                                                                                                                                                            \\ \hline
SYSTEM:                                     & Do you want to make a reservation for 2 people in the restaurant?                                                                                                                                                                                          \\ \hline
USER:                                       & Yes, thanks. What’s their phone number?                                                                                                                                                                                                                    \\ \hline
                                            & \begin{tabular}[c]{@{}l@{}}INTENT: ReserveRestaurant , a popular restaurant search and reservation service ,  make \\ a table reservation at a restaurant , SLOT: {\color[HTML]{00D2CB} number\_of\_seats} , number of seats to reserve at the restaurant\end{tabular}            \\ \cdashline{2-2} 
                                            & \begin{tabular}[c]{@{}l@{}}INTENT: ReserveRestaurant ,  a popular restaurant search and reservation service ,  make \\ a table reservation at a restaurant ,  SLOT: time , tentative time of restaurant reservation\end{tabular}                           \\ \cdashline{2-2} 
                                            & \begin{tabular}[c]{@{}l@{}}INTENT: ReserveRestaurant ,  a popular restaurant search and reservation service ,  make \\ a table reservation at a restaurant , SLOT: date , tentative date of restaurant reservation\end{tabular}                            \\ \cdashline{2-2} 
                                            & \begin{tabular}[c]{@{}l@{}}INTENT: ReserveRestaurant ,  a popular restaurant search and reservation service ,  make \\ a table reservation at a restaurant , SLOT: restaurant\_name,  name of the restaurant\end{tabular}                                  \\ \cdashline{2-2} 
\multirow{-5}{*}{\begin{tabular}[c]{@{}l@{}}RETRIEVAL: \\ (Predicted Top-5)\end{tabular}} & \begin{tabular}[c]{@{}l@{}}INTENT: ReserveRestaurant, a popular restaurant search and reservation service ,  make \\ a table reservation at a restaurant , SLOT: {\color[HTML]{00D2CB} location} , city where the restaurant is located\end{tabular}                              \\ \hline
APICALL: (Gold)                             & \begin{tabular}[c]{@{}l@{}}api\_name = ReserveRestaurant ; date = 2019-03-01 ; {\color[HTML]{00D2CB} location} = San Jose ; {\color[HTML]{00D2CB} number\_of\_seats} = 2 ; \\ restaurant\_name = Sino ;  time = 11:30\end{tabular}                                                                       \\ \hline
APICALL: (Predicted)                        & \begin{tabular}[c]{@{}l@{}}api\_name = ReserveRestaurant ; date = 2019-03-01 ; {\color[HTML]{CB0000} city} = San Jose ; {\color[HTML]{CB0000} party\_size} = 2 ; \\ restaurant\_name = Sino ; time = 11:30\end{tabular}                                                                                 \\ \hline
APIRESP:                                    & \begin{tabular}[c]{@{}l@{}}city = San Jose ; cuisine = Asian ; has\_live\_music = False ; phone\_number = 408-247-8880 ;  \\ price\_range = moderate ;  restaurant\_name: Sino;  serves\_alcohol = False ;  street\_address = 377 Santana Row\end{tabular} \\ \hline
SYSTEM:                                     & The phone number is 408-247-8880.                                                                                                                                                                                                                          \\ \hline
\end{tabular}} \caption{A predicted example on the development set. Red colors indicate incorrect predictions and light blue colors indicate correct slots. } \label{tab:error_analysis}
 \vspace{-1.1em}
\end{table*}

\paragraph{Influence of Irrelevant Information on the End-to-End TOD}
Given the potential emergence of unseen intents and slots in real-world scenarios, it is challenging to expect a perfect retrieval module. 

In this section, we first investigate the ability of the TOD to ignore irrelevant retrieved information.  In this section, we first investigate ``\textit{the TOD's ability in learning to ignore irrelevant retrieved information}''. 
Table \ref{tab:tod-ablation} shows the corresponding results. As described in Sec \ref{section-cache}, API-information includes all intents and slots for the whole dialogue.
The retrieval module exhibits an $84.4\%$ Top-1 accuracy in retrieving all slot information in a single attempt, as shown in Table \ref{tab:DPR}.
However, when setting $N$ to 5, the retriever returns similar yet irrelevant information despite a near $100\%$ Top-5 recall accuracy. This results in the inclusion of lots of irrelevant intents and slots into the generative model. Interestingly, ``\textit{API-information (N=1)}''  performs similar to ``\textit{API-information (N=5)}'', suggesting that the TOD is capable of learning to ignore irrelevant retrieved information.

Second, we investigate ``\textit{if the TOD generator relies more on the retriever when  all  retrieved information entries are stacked together}''. In pursuit of this objective, we compare FiD-TOD and FiD-TOD-NoStack, with the difference being whether the retrieved information entries are handled collectively or separately. As shown in Row 3 of Table~\ref{tab:tod-ablation}, FiD-TOD-NoStack performs slightly worse when not stacking all retrieved information directly with a single dialogue context. This could be attributed to the design of FiD-TOD-NoStack, which results in repeated dialogue context during each retrieval attempt and may hinder the retrieved information.

\paragraph{Error Analysis}
Despite the retrieval module demonstrating relatively high Top-5 accuracy, there is still room for improvement in the Joint Goal Accuracy (JGA).  Therefore, we examine potential reasons for this discrepancy.
Table \ref{tab:error_analysis} shows one most frequently appeared error type, where the retrieval module successfully retrieve Top-5 information entries from the cache. In terms of APICALL prediction, the TOD accurately generates the intent and associated values. Among the generated slots,``\textit{city}'' and ``\textit{party\_size}''  are semantically similar to  ``\textit{location}'' and ``\textit{number\_of\_seats}'', respectively. However, the two generated slots are incorrect as they belongs to different services. Upon further inspection, we find these terms are from the training cache. This suggests that the TOD generator does not completely rely on the retriever, and it tends to memorize the training slot information entries from the training cache, pointing towards the need for better generalized abilities. 
Furthermore, approximately $20\%$ dialogue turns on the developments set shows this issue, suggesting a huge space to  improve the performance.  We hypothesize that data augmentation, such as entity replacements in dialogue history,
could be one possible way to mitigate this problem. 
We leave further exploration of this issue to future work. 


\section{Conclusion}
This paper aims to enhance the performance of end-to-end TOD systems by incorporating a simple cache.
We begin by constructing a simple cache containing intents and slots. Subsequently, we fine-tune a retrieval module to extract the most relevant information entries.
Next, we train the end-to-end TOD model, enabling it to reference and ground both the dialogue history and the retrieved information during TOD generation.
Experimental results, based on a large-scale SGD dataset, demonstrate that our approach outperforms strong baselines.


\section{Acknowledgements}
We extend our gratitude to Moya Chen, Paul A. Crook, and Hu Xu for their insightful discussions. Additionally, we appreciate the valuable feedback provided by our anonymous reviewers.

\clearpage
\newpage
\bibliographystyle{acl_natbib}
\bibliography{acl2022}


\end{document}


\maketitle

\setcounter{table}{5}
\setcounter{figure}{4}

\appendix


\section{Additional Notes on the Use of NLI}
\label{app:nli}

There are other tasks modeling relationships between sentences.
Paraphrase~\citep{paranmt} and semantic relatedness~\citep{semeval} tasks are such examples.
It is possible to automatically create large-scale paraphrase datasets by machine translation~\citep{ppdb}.
However, our task is not a paraphrasing task, and creating negative examples is crucial and non-trivial~\citep{selectional-preference}.
In contrast, as described above, the NLI setting comes with negative examples by nature.
The semantic relatedness (or textual similarity) task is considered as a coarse-grained task compared to NLI, as discussed in the previous work~\citep{jmt}, in that the task measures semantic or topical relatedness.
This is not ideal for the intent detection task, because we need to discriminate between topically similar utterances of different intents.
In summary, the NLI task well matches our objective, with access to large datasets.

\section{A Note on the Threshold Selection} \label{threshold-selection}
\label{app:threshold}
Our joint score ($\mathrm{Acc}_\mathrm{in} + R_\mathrm{oos}$) in Section~4.2 gives the same weight to the two metrics, $\mathrm{Acc}_\mathrm{in}$ and $R_\mathrm{oos}$, compared to other combined metrics like $(C_\mathrm{in}+C_\mathrm{oos})/(N_\mathrm{in}+N_\mathrm{oos})$.
Such a combined metric can put much more weight on the in-domain accuracy when $N_\mathrm{in}$ and $N_\mathrm{oos}$ are imbalanced; Table~2 shows such imbalance on the development set.
\citet{oos-intent} sacrificed the OOS recall a lot, and the trade-off with respect to the threshold selection was not discussed.

\section{Training Details} \label{training-details}

\begin{table*}[t]
\centering
\resizebox{1.0\linewidth}{!}{
\begin{tabular}{l|ccc|ccc}
\hline
           & \multicolumn{3}{c|}{\textbf{Single domain}}                 & \multicolumn{3}{c}{\textbf{All domains}}                 \\ \cline{2-7} 
           & \textbf{Learning rate} & \textbf{Epoch}    & \textbf{Run} & \textbf{Learning rate} & \textbf{Epoch} & \textbf{Run} \\ \hline
Classifier & \{1e-4, 2e-5, 5e-5\}   & \{15, 25, 35\}    & 10             & \{1e-4, 5e-5\}         & \{15, 25, 35\} & 5              \\
Emb-kNN    & \{1e-4, 2e-5, 3e-5\}   & \{7, 10, 20, 25, 35\} & 10             &   \{2e-5, 5e-5\}        &  \{3, 5, 7\}        & 5              \\
DNNC       & \{1e-5, 2e-5, 3e-5, 4e-5\}   & \{7, 10, 15\}     & 10             & \{2e-5, 5e-5\}         & \{3, 5, 7\}   & 5              \\ \hline
\end{tabular}}\caption{some hyper-parameter settings for a few models.}\label{table:hyper-paramter}
\end{table*}

\begin{table*}[t]
\centering
\begin{tabular}{l|cc}
\hline
           & \textbf{5-shot}                 & \textbf{10-shot}                \\ \hline
Classifier & \{bs: 50, ep: 25.0, lr: 5e-05\} & \{bs: 50, ep: 35.0, lr: 5e-05\} \\ \hline
Emb-kNN    & \{bs: 200, ep: 7.0, lr: 2e-05\} & \{bs: 200, ep: 5.0, lr: 2e-05\}  \\ \hline
DNNC       & \{bs: 900, ep: 7.0, lr: 2e-05\} &  \{bs: 1800, ep: 5.0, lr: 2e-05\} \\ \hline
\end{tabular}
\caption{Best hyper-parameter settings for a few models on the all-domain experiments, where {\tt bs} is batch size, {\tt ep} represents epochs, {\tt lr} is learning rate.}\label{table:hyper-paramter-best-all}
\end{table*}

\paragraph{Dataset preparation}
To use the CLINC150 dataset~\citep{oos-intent}\footnote{\url{https://github.com/clinc/oos-eval}.} in our ways, especially for the single-domain experiments, we provide a zip file {\tt data\_preprocess\_for\_emnlp2020.zip} accompanied with the paper submission.

\paragraph{General training}\label{appendix-general-training}
This section describes the details about the model training in Section~4.3.
For each component related to RoBERTa and SRoBERTa, we solely follow the two libraries, transformers and sentence-transformers, for the sake of easy reproduction of our experiments.\footnote{\url{https://github.com/huggingface/transformers} and \url{https://github.com/UKPLab/sentence-transformers}.}
The example code to train the NLI-style models is also available.\footnote{\url{https://github.com/huggingface/transformers/tree/master/examples/text-classification}.}
We use the {\tt roberta-base} configuration\footnote{\url{https://s3.amazonaws.com/models.huggingface.co/bert/roberta-base-config.json}.} for all the RoBERTa/SRoBERTa-based models in our experiments.
All the model parameters including the RoBERTa parameters are updated during all the fine-tuning processes, where we use the AdamW~\citep{adamw} optimizer with a weight decay coefficient of 0.01 for all the non-bias parameters.
We use a gradient clipping technique~\citep{clip} with a clipping value of 1.0, and also use a linear warmup learning-rate scheduling with a proportion of 0.1 with respect to the maximum number of training epochs. 

\paragraph{Pre-training on NLI tasks}\label{appendix-pre-training}
For the pre-training on NLI tasks, we fine-tune a {\tt roberta-base} model on three publicly available datasets, i.e., SNLI~\citep{snli}, MNLI~\citep{mnli}, and WNLI~\citep{wnli} from the GLUE benchmark~\citep{glue}.
The optimizer and gradient clipping follow the above configurations.
The number of training epochs is set to $4$; the batch size is set to $32$; the learning rate is set to $2e-5$.
We use a linear warmup learning-rate scheduling with a proportion of $0.06$ by following \citet{roberta}.
The evaluation results on the development sets are shown in Table~\ref{table-pretrain}, where the low accuracy of WNLI is mainly caused by the data size imbalance.
We note that these NLI scores are not comparable with existing NLI scores, because we converted the task to the binary classification task for our model transfer purpose.

\paragraph{Text pre-processing}
For all the RoBERTa-based models, we used the RoBERTa {\tt roberta-base}'s tokenizer provided in the transformers library.\footnote{\url{https://github.com/huggingface/transformers/blob/master/src/transformers/tokenization_roberta.py}.}
We did not perform any additional pre-processing in our experiments.

\paragraph{Hyper-parameter settings}\label{appendix-hyper-parameter}
Table~\ref{table:hyper-paramter} shows the hyper-parameters we tuned on the development sets in our RoBERTa-based experiments.
For a single-domain experiment, we take a hyper-parameter set and apply it to the ten different runs to select the threshold in Section~4.2 on the development set.
We then select the best hyper-parameter set along with the corresponding threshold, and finally apply the model and the threshold to the test set.
We follow the same process for the all-domain experiments, except that we run each experiment five times.
Table~\ref{table:hyper-paramter-best-all} and Table~\ref{table:hyper-paramter-best} summarize the hyper-parameter settings used for the evaluation on the test sets.
We note that each model was not very sensitive to the different hyper-parameter settings, as long as we have a large number of training iterations.

\begin{table}[]
\centering
\resizebox{\linewidth}{!}{
\begin{tabular}{l|lll}
\hline
Dataset                    & SNLI & WNLI & MNLI \\ \hline
Size of the development set & 9999          & 70            & 9814          \\
Accuracy                   & 94.5\%        & 41.4\%        & 92.1\%        \\ \hline
\end{tabular}}\caption{Development results on three NLI datasets.}
\label{table-pretrain}
\end{table}

\begin{table*}[t]
\centering
\resizebox{\linewidth}{!}{
\begin{tabular}{l|cccc}
\hline
           & \textbf{5-shot}                                      & \multicolumn{1}{c|}{\textbf{10-shot}}                 & \textbf{5-shot}                  & \textbf{10-shot}                 \\ \cline{2-5} 
           & \multicolumn{2}{c|}{\textbf{Banking}}                                                                                 & \multicolumn{2}{c}{\textbf{Credit cards}}                                    \\ \hline
Classifier & \multicolumn{1}{l}{\{bs: 15, ep: 25.0, lr: 5e-05\}}  & \multicolumn{1}{l|}{\{bs: 15, ep: 35.0, lr: 5e-05\}}  & \{bs: 15, ep: 15.0, lr: 5e-05\}  & \{bs: 15, ep: 25.0, lr: 5e-05\}  \\
Emb-kNN    & \multicolumn{1}{l}{\{bs: 200, ep: 35.0, lr: 1e-05\}} & \multicolumn{1}{l|}{\{bs: 200, ep: 25.0, lr: 2e-05\}} & \{bs: 100, ep: 20.0, lr: 1e-05\} & \{bs: 100, ep: 10.0, lr: 1e-05\} \\
DNNC       & \multicolumn{1}{l}{\{bs: 370, ep: 15.0, lr: 1e-05\}} & \multicolumn{1}{l|}{\{bs: 370, ep: 7.0, lr: 2e-05\}}  & \{bs: 370, ep: 15.0, lr: 2e-05\} & \{bs: 370, ep: 7.0, lr: 3e-05\}  \\ \hline
           & \multicolumn{2}{c}{\textbf{Work}}                                                                            & \multicolumn{2}{c}{\textbf{Travel}}                                 \\ \hline
Classifier & \{bs: 15, ep: 15.0, lr: 5e-05\}                      & \{bs: 15, ep: 15.0, lr: 5e-05\}                       & \{bs: 15, ep: 35.0, lr: 5e-05\}  & \{bs: 15, ep: 25.0, lr: 1e-04\}  \\
Emb-kNN    & \{bs: 100, ep: 20.0, lr: 1e-05\}                     & \{bs: 100, ep: 7.0, lr: 2e-05\}                       & \{bs: 100, ep: 35.0, lr: 3e-05\} & \{bs: 100, ep: 20.0, lr: 1e-05\} \\
DNNC       & \{bs: 370, ep: 7.0, lr: 3e-05\}                      & \{bs: 370, ep: 15.0, lr: 2e-05\}                      & \{bs: 370, ep: 7.0, lr: 2e-05\}  & \{bs: 370, ep: 7.0, lr: 2e-05\}  \\ \hline
\end{tabular}}\caption{Best hyper-parameter settings for a few models on the four single domains, where {\tt bs} is batch size, {\tt ep} represents epochs, {\tt lr} is learning rate.}\label{table:hyper-paramter-best}
\end{table*}

\section{Data Augmentation} \label{data-augmentation}

We describe the details about the classifier baselines with the data augmentation techniques in Section~4.3.

\paragraph{EDA}
Classifier-EDA uses the following four data augmentation techniques in \citet{eda}: synonym replacement, random insertion, random swap, and random deletion.
We follow the publicly available code.\footnote{\url{https://github.com/jasonwei20/eda_nlp}.}
For every training example, we empirically set one augmentation based on every technique.
We apply each technique separately to the original sentence and therefore every training example will have four augmentations.
The probability of a word in an utterance being edited is set to 0.1 for all the techniques.   

\paragraph{BT}
For classifier-BT, we use the English-German corpus in \citet{escape}, which is widely used in an annual competition for automatic post-editing research on IT-domain text~\citep{ape-2019}.
The corpus contains about 7.5 million translation pairs, and we follow the {\it base} configuration to train a transformer model~\citep{transformer} for each direction.
Based on the initial trial in our preliminary experiments to generate diverse examples, we decided to use a temperature sampling technique instead of a greedy or beam-search strategy.
More specifically, logit vectors during the machine translation process are multiplied by $\tau$ to distort the output distributions, where we set $\tau = 5.0$.
For each training example in the intent detection dataset, we first translate it into German and then translate it back to English.
We repeat this process to generate up to five unique examples, and use them to train the classifier model.
Table~\ref{tb:bt-examples} shows such examples, and we will release all the augmented examples for future research.

\begin{table*}[t]
  \begin{center}
{\small
    \begin{tabular}{l|l|l}
    
    Original utterance & Augmented example & Intent label \\ \hline
    can you block my chase account right away please & can you turn my chase account off directly & freeze account \\
    do a car payment from my savings account & with my saving account, you can pay a car payment account & pay bill \\
    when is my visa due & when is my visa to be paid & bill due \\ \hline

    \end{tabular}
}
    \caption{Examples used to train clasifier-BT.}
    \label{tb:bt-examples}
  \end{center}

\end{table*}

\section{More Results}
\label{extra-results}

\paragraph{Visualization} \label{appendix-vidualization}
Figure~\ref{fig:visulization-appendix} shows the same curves in Figure~3 along with the corresponding 10-shot results.
We can see that the 10-shot results also exhibit the same trend.
Figure~\ref{fig:tsne-appendix} shows more visualization results with respect to Figure~1.
Again, the 10-shot visualization shows the same trend.

Figure~\ref{fig:Conf-appendix} and Figure \ref{fig:Conf-appendix-all-domains} show 5-shot and 10-shot confidence levels on the test sets of the banking domain and all domains, respectively.
Both Classifier and Emb-kNN cannot perform well to distinguish the in-domain examples from the OOS examples, while DNNC has a clearer distinction between the two.

\paragraph{Faster inference}\label{appendix-DNNC-joint}
Figure~\ref{fig:joint_nli-appendix} shows the same curves in Figure~4 also for the 10-shot setting.
We can see the same trend with the 10-shot results.

\paragraph{Case studies}\label{Case Study}
Table~\ref{table:case-study} shows four DNNC prediction examples from the development set of the banking domain.
For the first example, the input utterance is correctly predicted with a high confidence score, and it has a similarly matched utterance to the input utterance;
for the second example, the input utterance is predicted incorrectly with a high confidence score, where the matched utterance is related to money but it has a slightly different meaning with the input utterance.
For the third example, the model gives a very low confidence score to predict an OOS user utterance as an in-domain intent; the last example is an incorrect case where the input utterance and the matched utterance have a topically similar meaning, resulting in a high confidence score for the wrong label, ``bill due.''
Based on these observations, it is an important direction to improve the model's robustness (even with the large-scale pre-trained models) towards such confusing cases.

\begin{figure*}[t]
	\begin{center}
    	\includegraphics[width=0.85\linewidth]{./images_appendix/Visualization_Metric_Appendix.png}
    \end{center}
\caption{5-shot and 10-shot development results on the banking domain. In this series of plots, a model with a higher area-under-the-curve is more robust.}
\label{fig:visulization-appendix}
\end{figure*}

\begin{figure*}[t]
	\begin{center}
    	\includegraphics[width=0.85\linewidth,height=0.9\textheight]{./images_appendix/TSNE_Appendix.png}
    \end{center}
\caption{5-shot and 10-shot tSNE visualizations on development set of the banking domain, where circles represent in-domain intent classes, and red stars represent out-of-scope intents.}
\label{fig:tsne-appendix}
\end{figure*}

\begin{figure*}[t]
	\begin{center}
    	\includegraphics[width=0.85\linewidth,keepaspectratio=true,height=0.95\textheight]{./images_appendix/Conf_Appendix_with_sbert.png}
    \end{center}
\caption{5-shot and 10-shot confidence levels on test set of the banking domain. Best viewed in color.}
\label{fig:Conf-appendix}
\end{figure*}

\begin{figure*}[t]
	\begin{center}
    	\includegraphics[width=0.85\linewidth]{./images_appendix/Conf_Appendix_All_domains.png}
    \end{center}
\caption{5-shot and 10-shot confidence levels on test set of all domains. Best viewed in color.}
\label{fig:Conf-appendix-all-domains}
\end{figure*}

\begin{figure*}[t]
	\begin{center}
    	\includegraphics[width=0.85\linewidth]{./images_appendix/Visualization_Joint_nli_Appendix.png}
    \end{center}
\caption{5-shot and 10-shot DNNC-joint development results on the banking domain, where the dash lines are DNNC results.}
\label{fig:joint_nli-appendix}
\end{figure*}

\begin{table*}[]
\centering
\resizebox{1.0\linewidth}{!}{
\begin{tabular}{ll}
\hline
\textbf{input utterance}    & transfer ten dollars from my wells fargo account to my bank of america account              \\
\textbf{matched utterance} & transfer \$10 from checking to savings                                                      \\
\textbf{label of the input utterance}      & transfer                                                                                    \\
\textbf{label of the matched utterance}      & transfer                                                                                    \\
\textbf{confidence score}             & 0.934                                                                                      \\ \hline
\textbf{input utterance}    & what transactions have i accrued buying dog food                                            \\
\textbf{matched utterance} & what have i spent on food recently                                                          \\
\textbf{label of the input utterance}      & transactions                                                                                \\
\textbf{label of the matched utterance}      & spending history                                                                           \\
\textbf{confidence score}             & 0.915                                                                                      \\ \hline
\textbf{input utterance}    & who has the best record in the nfl                                                          \\
\textbf{matched utterance} & do i have enough in my boa account for a new pair of skis                                   \\
\textbf{label of the input utterance}      & OOS                                                                                         \\
\textbf{label of the matched utterance}      & balance                                                                                     \\
\textbf{confidence score}             & 0.006                                                                                       \\ \hline
\textbf{input utterance}    & how long will it take me to pay off my card if i pay an extra \$50 a month over the minimum \\
\textbf{matched utterance} & how long do i have left to pay for my chase credit card                                     \\
\textbf{label of the input utterance}      & OOS                                                                                         \\
\textbf{label of the matched utterance}      & bill due                                                                                   \\
\textbf{confidence score}             & 0.945                                                                                      \\ \hline
\end{tabular}} \caption{Case studies on the development set of banking domain. The first two cases are in-domain examples from the banking domain, and the rest are OOS examples.}\label{table:case-study}
\end{table*}

\newpage
\bibliographystyle{acl_natbib}
\bibliography{emnlp2020}